\documentclass[conference]{IEEEtran}
\IEEEoverridecommandlockouts
\usepackage{cite}
\usepackage{amsmath,amssymb,amsfonts}
\usepackage{algorithmic}
\usepackage{graphicx}
\usepackage{textcomp}
\usepackage{xcolor}

\usepackage[T1]{fontenc}

\usepackage{subcaption}
\definecolor{cvprblue}{RGB}{97, 153, 201}  
\usepackage[colorlinks=true, citecolor=cvprblue]{hyperref} 

\def\BibTeX{{\rm B\kern-.05em{\sc i\kern-.025em b}\kern-.08em
    T\kern-.1667em\lower.7ex\hbox{E}\kern-.125emX}}
\begin{document}

\title{A Novel Retinal Image Contrast Enhancement – Fuzzy-Based Method}

\author{
    \IEEEauthorblockN{Adnan Shaout\textsuperscript{*}} 
    \IEEEauthorblockA{
        \textit{Department of Engineering and Computer Science} \\ 
        \textit{University of Michigan-Dearborn} \\ 
        Dearborn, Michigan, USA \\ 
        Email: shaout@umich.edu
    }
    \and
    \IEEEauthorblockN{Jiho Han\textsuperscript{*}} 
    \IEEEauthorblockA{
        \textit{Department of Engineering and Computer Science} \\ 
        \textit{University of Michigan-Dearborn} \\ 
        Dearborn, Michigan, USA \\ 
        Email: jihohan@umich.edu
    }
    \thanks{\textsuperscript{*}Joint First Authors.}
}

\maketitle
\begin{abstract}
\label{sec:abstract}
The vascular structure in retinal images plays a crucial role in ophthalmic diagnostics, and its accuracies are directly influenced by the quality of the retinal image. Contrast enhancement is one of the crucial steps in any segmentation algorithm – more so since the retinal images are related to medical diagnosis. Contrast enhancement is a vital step that not only intensifies the darkness of the blood vessels but also prevents minor capillaries from being disregarded during the process. This paper proposes a novel model that utilizes the linear blending of Fuzzy Contrast Enhancement (FCE) and Contrast Limited Adaptive Histogram Equalization (CLAHE) to enhance the retinal image for retinal vascular structure segmentation. The scheme is tested using the Digital Retinal Images for Vessel Extraction (DRIVE) dataset. The assertion was then evaluated through performance comparison among other methodologies which are Gray-scaling, Histogram Equalization (HE), FCE, and CLAHE.  It was evident in this paper that the combination of FCE and CLAHE methods showed major improvement. Both FCE and CLAHE methods, dominating with 88\% as better enhancement methods, proved that preprocessing through fuzzy logic is effective. 
\end{abstract}

\section{Introduction}
Countless ocular diseases, including but not limited to diabetic retinopathy, melanoma, age-related macular degeneration, etc., beget ophthalmic sequelae to the patients, leaving them with severe impairments within their daily lives to overcome \cite{HowRetinalImaging}. Such threats can be easily avoided by detecting the disease in the early stages.  Retinal scanning is one of many tests that doctors can do to achieve this. Clear images of the retina allow ophthalmologists to detect the patients’ eye health, and contrast enhancement plays a crucial role in aiding the ophthalmologists in making accurate identifications.

In this paper, we propose a contrast enhancement method that leverages the strengths of both fuzzy logic and CLAHE. While fuzzy logic enhances contrast effectively, it struggles with sporadic bright and dark spots that appear across the image. On the other hand, we figured that the CLAHE method excels at segmenting blood vessels correctly in such outliers but experiences difficulties creating crisp distinctions between tissues and capillaries during segmentation due to its low contrast. Therefore, we sought a novel scheme that can utilize the advantages of both methods through linear blending combined with post-processing of the image.

Methods of retinal imaging may vary, but the basics – lights pass through from the pupil to the retina, and the machine retrieves the collected image – remain unchanged. However, such method results in the variance of luminosity throughout the image and noises within the resultant image, which is the main challenge of contrast enhancement in retinal imaging.
According to an article by Naveed et al. on Diagnostics in 2021, poor image contrast and quality result from the complex image-collecting process, and unwanted variables corrupt the image, including, but not limited to, additive noise (Gaussian noise), multiplicative noise (speckle), and Shot noise (Poisson noise) \cite{naveedAutomatedEyeDiagnosis2021}.

Medical images are mostly in grayscale so that doctors can get a clear and better perception of images. This is because gray scaling can reduce the 3-dimensional pixel value consisting of red, blue, and green (RGB) into a 1-dimensional value \cite{youssefFuzzybasedImageSegmentation2021}. However, research proved that the green channel (G-channel) of RGB provides the best contrast of the blood vessels \cite{GreenChannelOverview}. Moreover, studies on the sensitivity functions of the human eye in color scales have proven that both the low-contrast signal detectability improved significantly while not affecting eye fatigue with a yellow-scale display over the grayscale display \cite{oguraComparisonGrayscaleColorscale2017}.

Innovative attempts have been made to improve the accuracy of retinal vessel segmentation. Some used fuzzy logic for edge detection in Diabetic Retinopathy, whose image shows lots of unwanted noise \cite{kamilEdgeDetectionDiabetic2014}. Recent methods rely much on machine learning (\!\!\cite{girdharRetinalImageSegmentation2022, barrosMachineLearningApplied2020}), convolutional neural networks (\!\!\cite{hoqueDeepLearningRetinal2021, khanalDynamicDeepNetworks2020}), and deep learning \cite{shiDeepLearningSystem2022}. Even though they perform very well, such methods can get very complex. We are attempting to match their performance while being less complex. Various methods and their accuracy can be found in \cite{PapersCodeDRIVE}

The paper is organized as follows: Section \ref{sec:methodologies} introduces the proposed model, Section \ref{sec:implementation} presents the implementation of the proposed method, and Section \ref{sec:results} presents results and conclusions.
\section{Proposed Model and its Methodology}
\label{sec:methodologies}

\begin{figure}[t]
    \centering
    \includegraphics[width=0.9\linewidth]{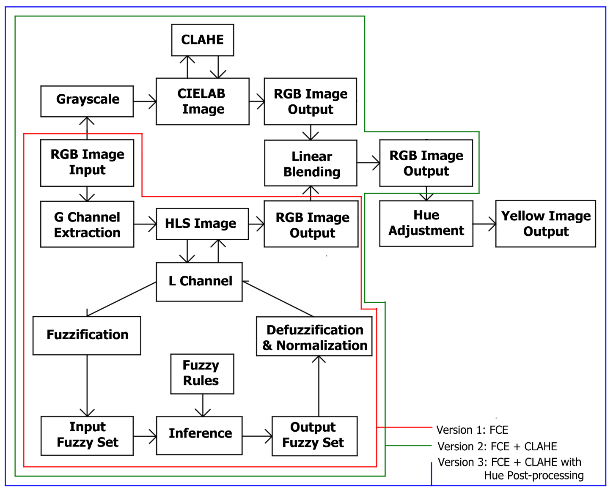}
    \caption{Block diagram of the proposed model.}
    \label{fig:block diagram}
\end{figure}

Figure \ref{fig:block diagram} shows the block diagram for the proposed model.  For this paper, the DRIVE Dataset has been used to test and create the outputs of the model. Three versions of the proposed model have been put to the test, each attempting to improve upon the former version in a multitude of ways. The versions are as follows:

\subsection{Fuzzy Contrast Enhancement (Version 1)}
\label{subsec:version_1}
The method for Fuzzy Contrast Enhancement is inspired by the fuzzy-based image enhancement method proposed by V\^{u}ong\ L\^e\ Minh\ Nguy\~en
\cite{FuzzyLogicImage}. The method is as follows:

\subsubsection{Import RGB Image and G-channel Extraction}
\begin{figure}[b]
    \centering
    \includegraphics[width=0.9\linewidth]{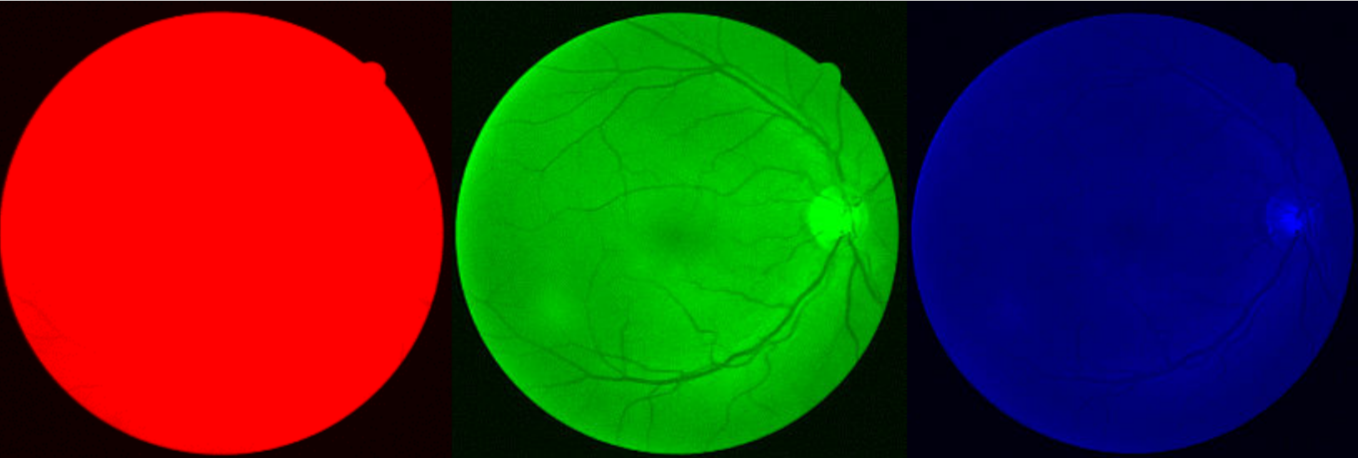}
    \caption{Comparison of the R, G, and B channels of the same fundus image.}
    \label{fig:RGB comparison}
\end{figure}

First, the image gets imported from the designated directory in an RGB format, and the G-channel of the image is extracted to reduce the size of the image.  OpenCV was used for implementing both functionalities.  Figure \ref{fig:RGB comparison} shows a comparison of the R, G, and B channels of the same image.

\subsubsection{Conversion to Hue-Luminosity-Saturation (HLS) format and L-Channel Extraction}
he image is converted from RGB format into Hue-Luminosity-Saturation (HLS) color space format. The International Commission on Illumination $L^{*}a^{*}b^{*}$ color-space (CIELAB) format is more widely used in modifying the luminosity/lightness of the image due to its design putting the relative perceptual difference of the human eye into consideration. However, as only a single color channel is being used for this step, using the HLS format was more efficient, as both hue and saturation of HLS color space and $a^{*}$ and $b^{*}$ of CIELAB are static in the same color channel, and conversion from RGB to CIELAB costs more computational time in comparison to that of RGB to HLS.

\subsubsection{Fuzzy Transform of Luminosity in L Channel Image}
In this part, the luminosity ranging from 0 to 100 is fuzzified into five linguistic values (very dark, dark, medium, bright, and very bright) using the Gaussian membership function.  Figure 3 shows the graphic analysis of the membership functions in different mean luminosity.  The mathematical equations for the five linguistic memberships are as follows:

\begin{align}
    \mu_{\text{VeryDark}} =& G \left( x, \max(-20, M - 40), \frac{M}{2} \right) \notag \\
    \mu_{\text{Dark}} =& G \left( x, 0.45M, \frac{M}{4} \right) \notag \\
    \mu_{\text{Medium}} =& G \left( x, 1.1M, \frac{M}{6} \right) \notag \\
    \mu_{\text{Bright}} =& G \left( x, 2.5M - 25, \frac{100 - M}{4} \right) \notag \\
    \mu_{\text{VeryBright}} =& G \left( x, 125, \frac{100 - M}{4} \right) \notag \\
    \text{where} \quad & G(x, M, \sigma) = e^{-0.5 \left( \frac{x - M}{\sigma} \right)^2}
    \label{eq:contrast_enhancement}
\end{align}

\begin{figure}[h]
    \centering
    \includegraphics[width=\linewidth]{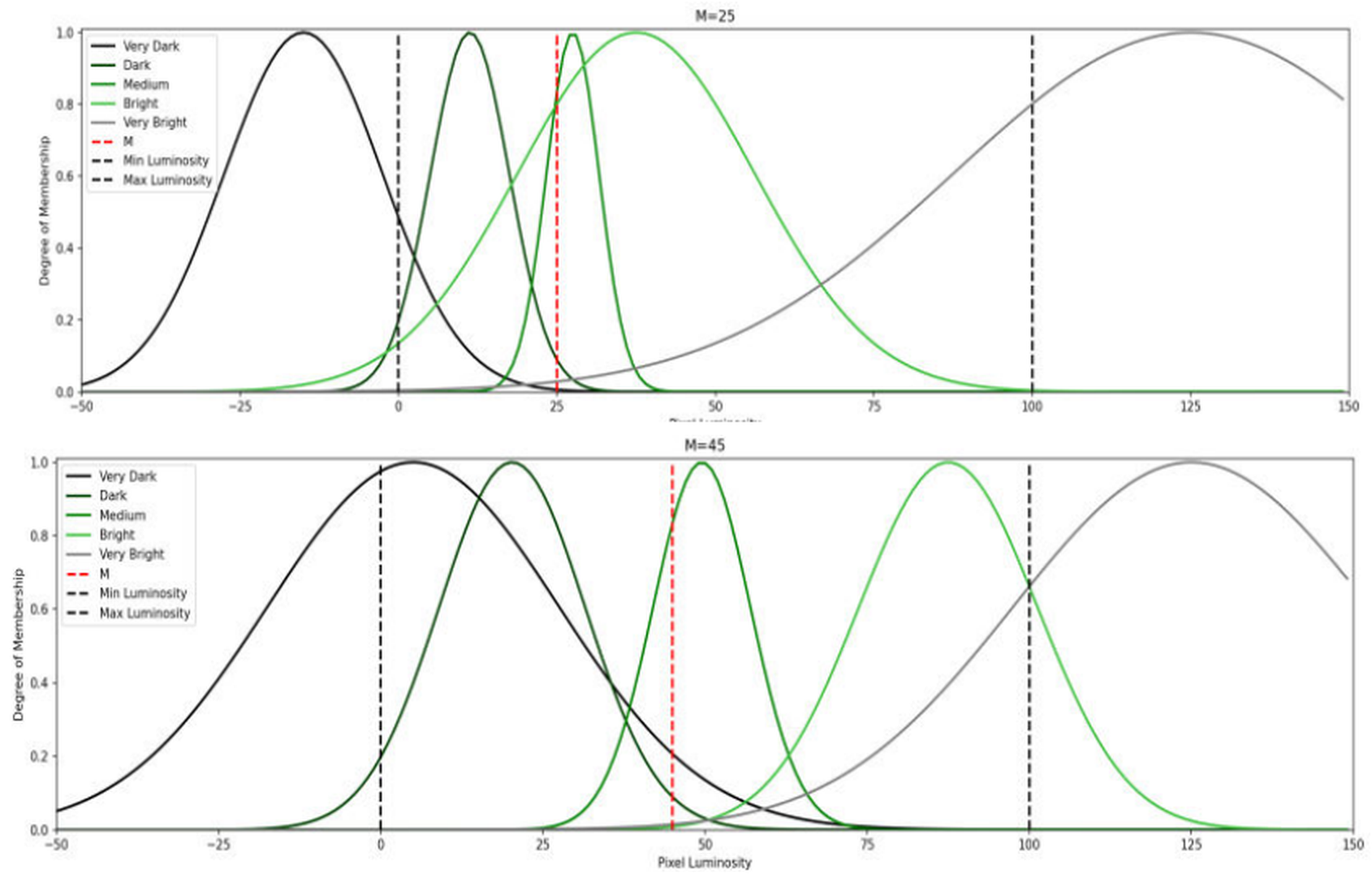}
    \caption{Visualization of the membership functions in different mean luminosity conditions.}
    \label{fig:adaptive_thresholding}
\end{figure}

Here, $x$ is the luminosity of the pixel, and $\sigma$ is the standard deviation of the luminosity of the entire image. $M$ is the variance-reduced mean luminosity of the whole image, and its value is calculated based on the mean luminosity of the image. Using the $M$ and $\sigma$ as input allows the membership function to generally shift around based on the overall brightness of the image itself, as the level of darkness of blood vessels compared to tissues is subjective. The Gaussian function was used as a membership function, and it is one of the most popular membership functions due to its conciseness and smoothness. The adaptive shift of the threshold via Eq. (\ref{eq:contrast_enhancement}) is visualized in Fig. \ref{fig:adaptive_thresholding}.

Three fuzzy rules were used for this version as follows:
\begin{itemize}
    \item{IF input is Dark THEN output is Very Dark}
    \item{IF input is Medium THEN output is Medium}
    \item{IF input is Bright THEN output is Very Bright}
\end{itemize}

This process intensifies the contrast of the image, i.e., making the bright part brighter and the dark part darker.

\subsubsection{Normalization by Min-max scaling of new Luminosity}
Passing through such rules may result in luminosity values varying beyond their original range of 0 to 100 or luminosity values not varying enough. To remedy such issues, the luminosity is normalized so that the luminosity values range fully from 0 to 100 using the following equation:
\begin{equation}
    L_{\text{norm}} = \left( \frac{L - L_{\min}}{L_{\max} - L_{\min}} \right) \times 100
    \label{eq:lnorm}
\end{equation}

\subsubsection{Output and Its Limitations}
The final output image, produced after applying Fuzzy Contrast Enhancement (FCE), exhibits improved contrast, enhancing the visibility of the retinal blood vessels. However, this method alone presents some limitations in preserving finer vascular details in high-luminosity regions.

\begin{figure}[h]
    \centering
    \includegraphics[width=0.7\linewidth]{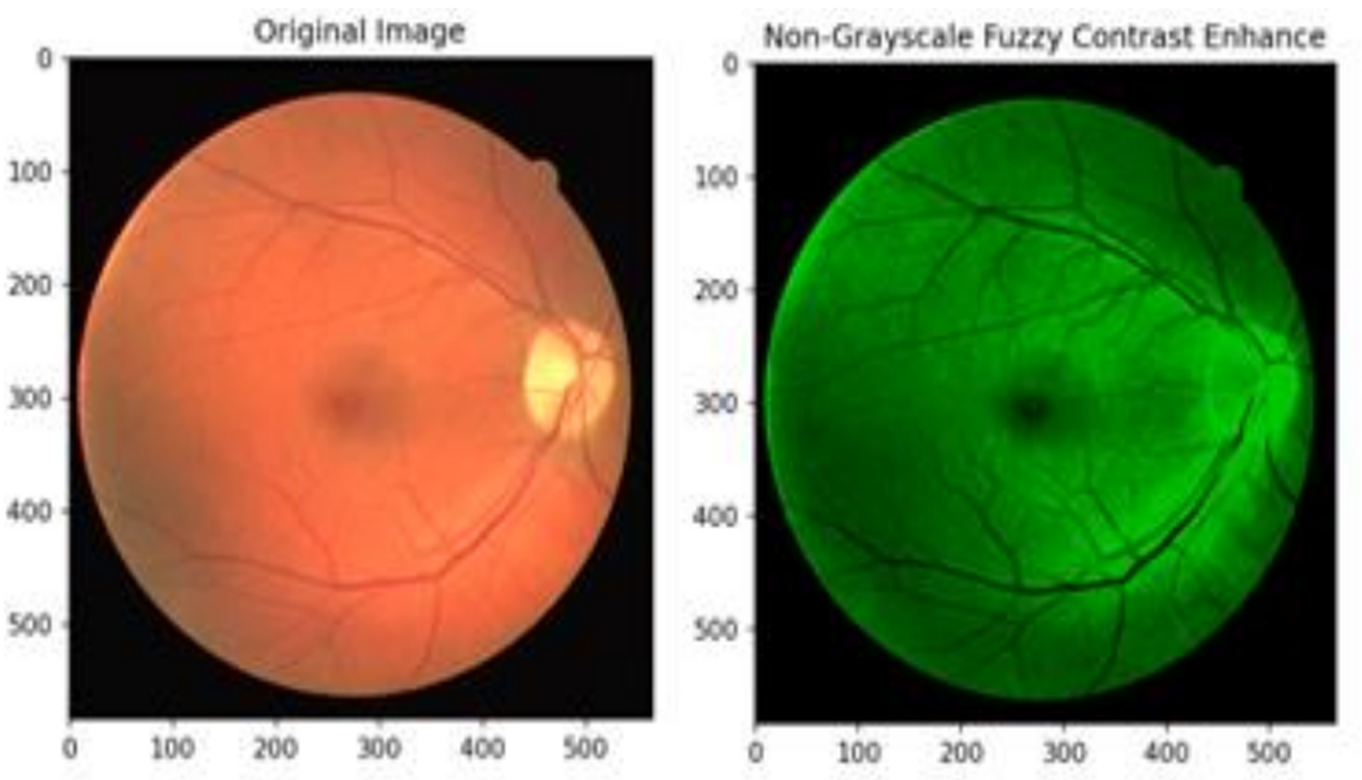}
    \caption{Comparison of the original image and FCE image, where the visibility of the vascular structure improved overall but deteriorated in the macular area.}
    \label{fig:FCE_results}
\end{figure}

Figure \ref{fig:FCE_results} illustrates the original image compared to the enhanced output, highlighting both the improvements in vessel contrast and the areas where the method struggles to preserve fine details.

\subsection{Post-processing}
\subsubsection{Linear blending (Version 2)}
\label{subsec:version_2}

The outputs of both methods, fuzzy and CLAHE, are linearly blended using the following equation:
\begin{equation}
    \text{output}_{\text{total}} = w_{1} \cdot \text{output}_{\text{fuzzy}} + w_{2} \cdot \text{output}_{\text{CLAHE}} - c
    \label{eq:output_total}
\end{equation}

where $w_{1}$, $w_{2}$ are the blending weights, and $c$ is a constant. In our implementation, we set $w_{1} = 0.6$, $w_{2} =  0.8$, and $c = -0.4$ based on empirical observations.

The selection of the weights and the constant was based on specific considerations. The weight of the CLAHE method is higher than that of the fuzzy method. The CLAHE method has, although not as crisp as FCE, shown better accuracy in increasing contrast between blood vessels and retinal tissues.  Ensuring accurate preprocessing of blood vessels is of paramount importance, as segmentation errors cannot be justified for marginal improvements in morphological representation. The constant was added not only to normalize the overall luminosity of the image but also to allow non-programmer personnel to edit the output luminosity with ease, as not all images share the same luminosity value.

\subsubsection{Hue adjustment for yellow output image (Version 3)}
\label{subsec:version_3}
The hue of the output of the image is finally modified so that the overall color of the image has a yellowish hue. This is because the yellowish hue performs the best in supporting doctors to better recognize distinctions within relatively low-contrast images \cite{oguraComparisonGrayscaleColorscale2017}.
\section{Implementation}
\label{sec:implementation}
The overall software is written in Python with the support of multiple libraries.

\subsection{Version 1 (FCE)}
\begin{figure}[h]
    \centering
    \includegraphics[width=0.7\linewidth]{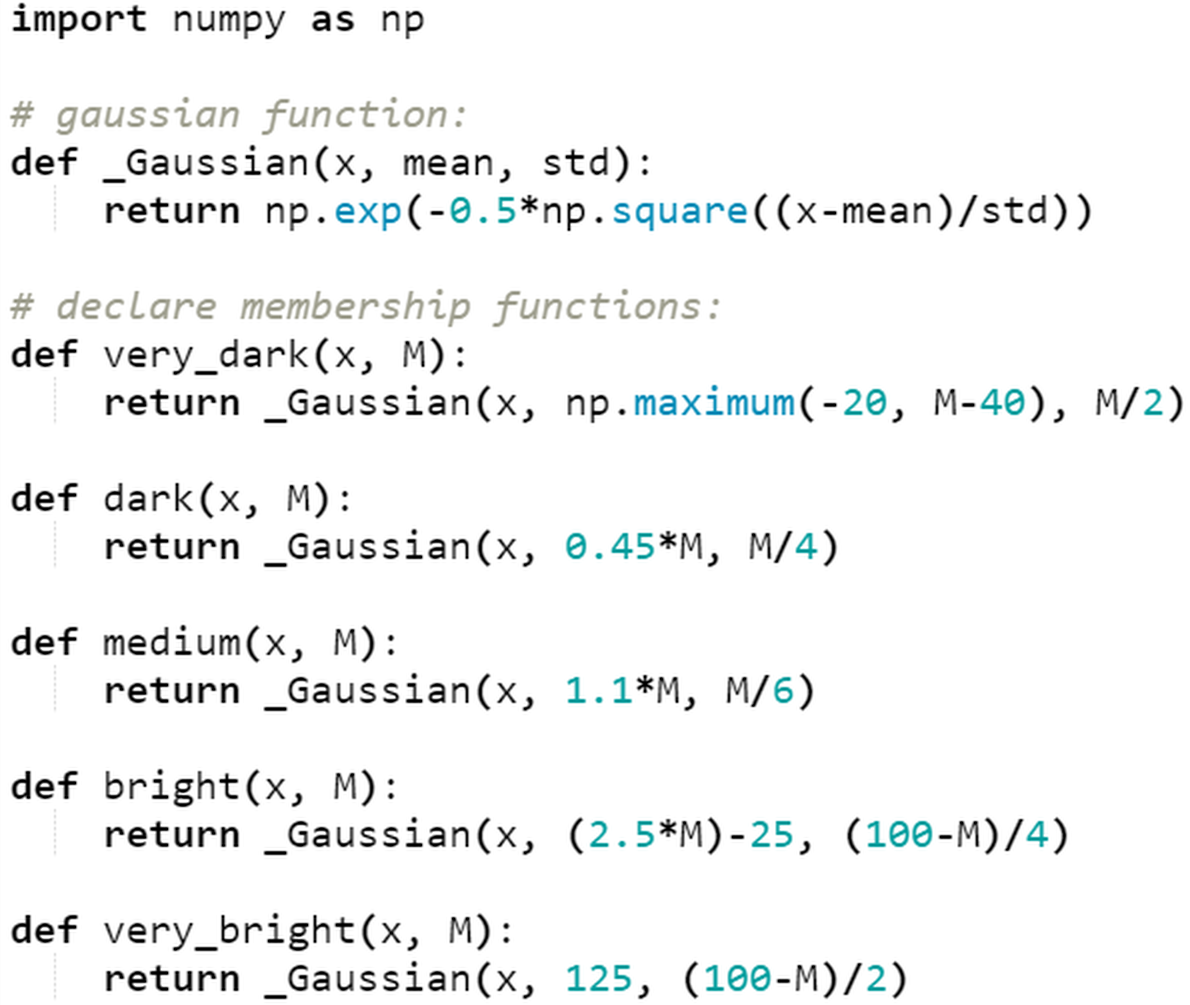}
    \caption{Implementation of rules for membership functions.}
    \label{fig:rules}
\end{figure}

Images are loaded and altered through OpenCV’s functions. Fuzzy rules are implemented through separate Python functions, as shown in Fig. \ref{fig:rules}.

\begin{figure}[h]
    \centering
    \includegraphics[width=\linewidth]{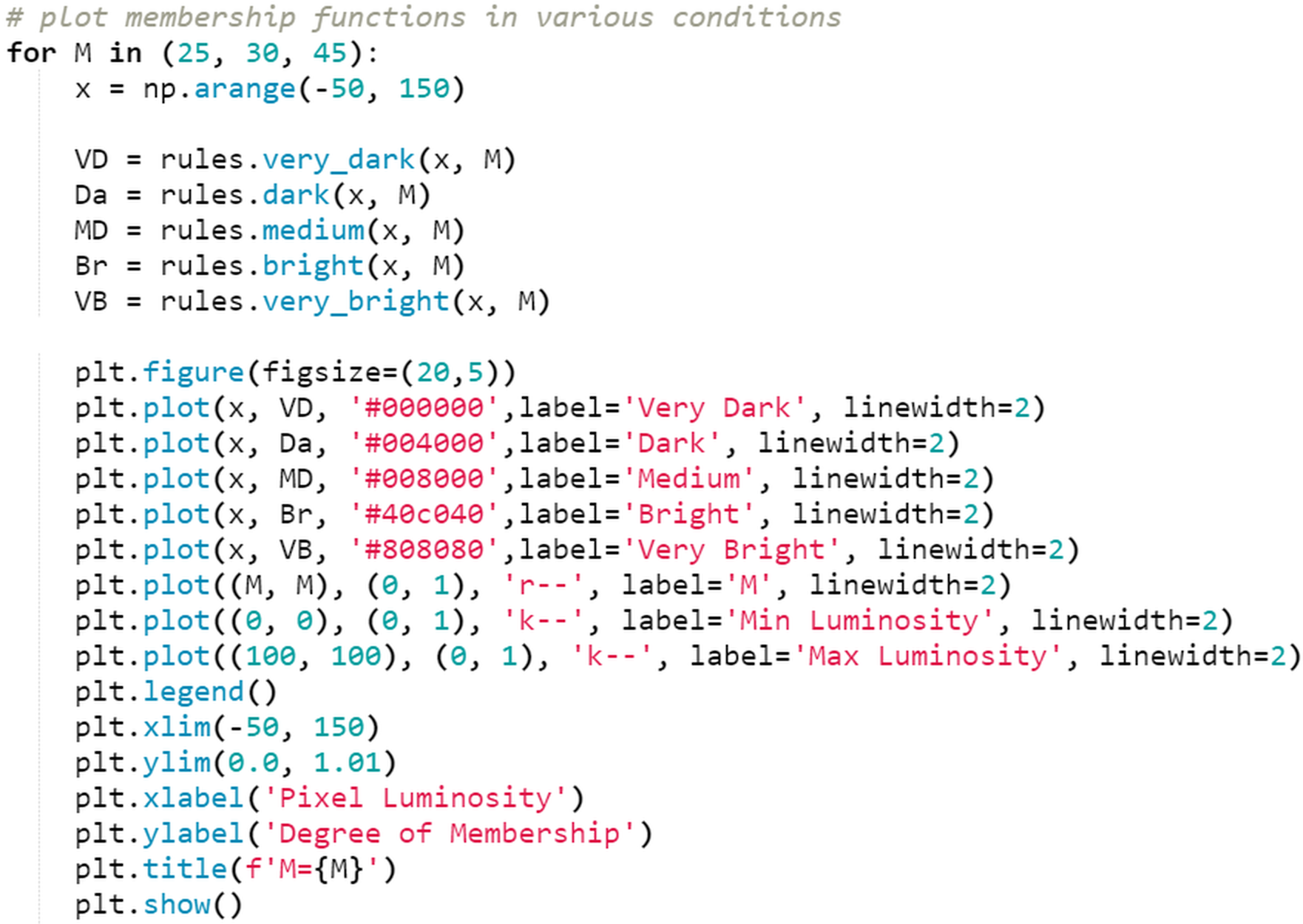}
    \caption{Visualization of fuzzy membership functions.}
    \label{fig:visusalization}
\end{figure}

The plotting of these membership functions for various values is done through matplotlib, as shown in Fig. \ref{fig:visusalization}.

\begin{figure}[h]
    \centering
    \includegraphics[width=\linewidth]{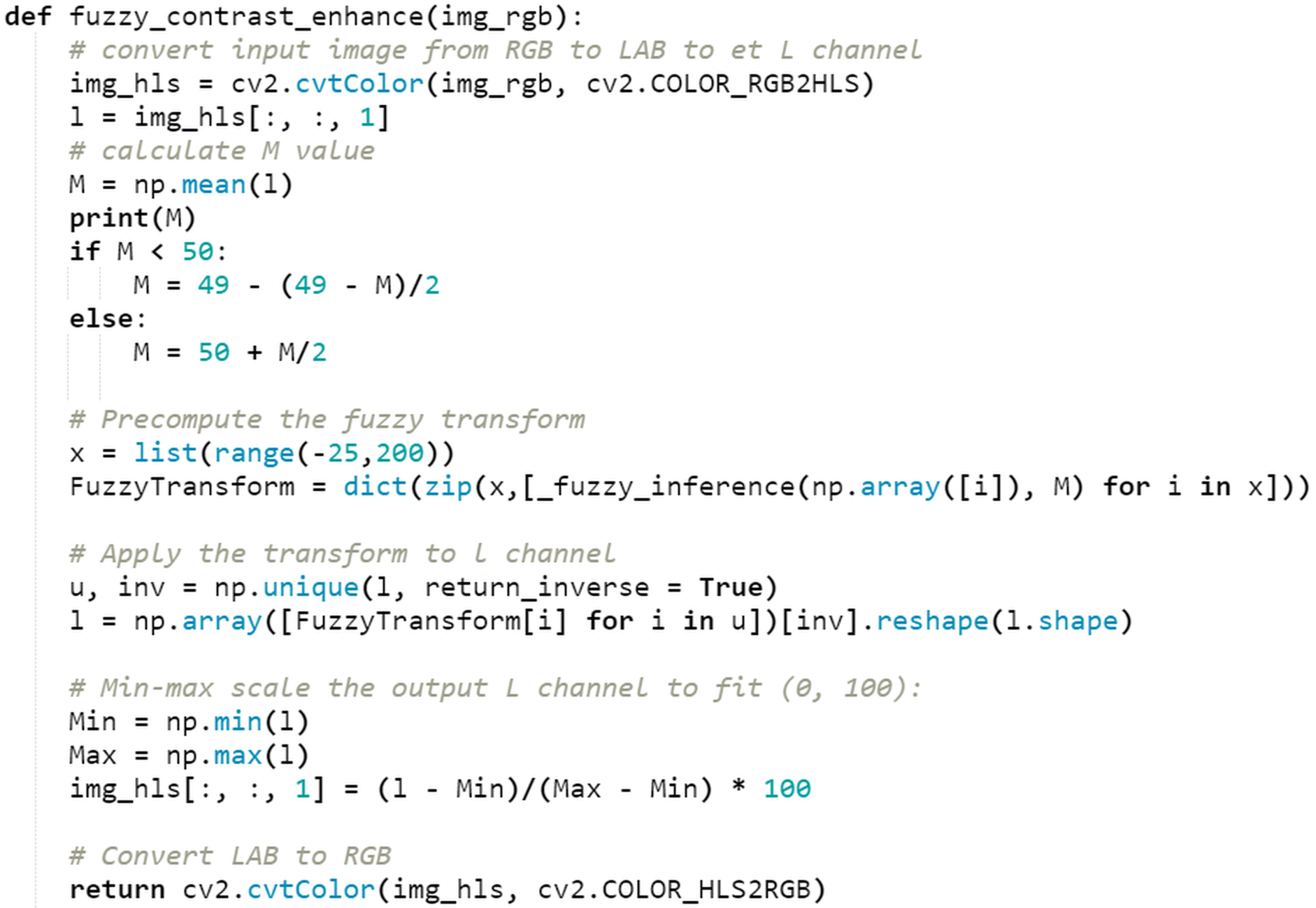}
    \caption{Implementation of FCE.}
    \label{fig:FCE}
\end{figure}

Fuzzy Contrast Enhance is done through fuzzification, inference, and defuzzification, which was done through sets of functions as shown in Fig. \ref{fig:FCE}. The input RGB image, denoted as \texttt{img\_rgb}, is converted to the HLS color space. Amongst the HLS color space, the L channel is extracted, and its variance-reduced mean value is calculated and stored as \texttt{M}. The fuzzy transform is performed onto the L channel using the saved \texttt{M} value, and min-max scaling has been performed.

\subsection{Version 2 (FCE + CLAHE)}
The functionality of CLAHE is implemented using OpenCV’s pre-existing function – \texttt{createCLAHE()}.

\begin{figure}[h]
    \centering
    \includegraphics[width=\linewidth]{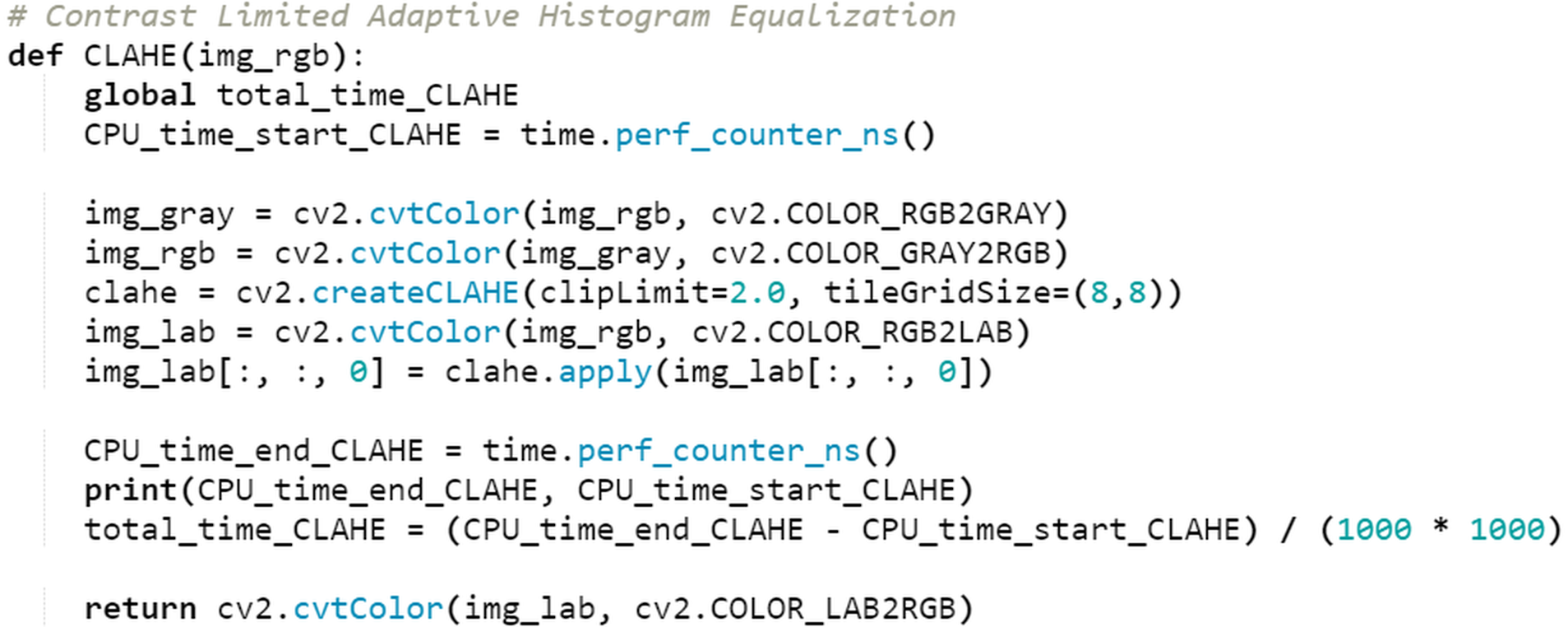}
    \caption{Implementation of CLAHE.}
    \label{fig:CLAHE}
\end{figure}

\begin{figure}[h]
    \centering
    \includegraphics[width=0.9\linewidth]{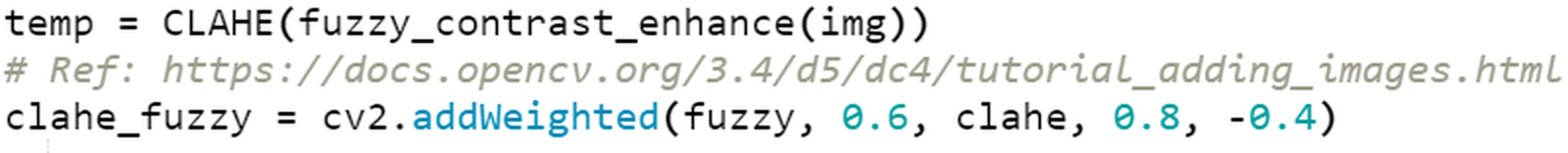}
    \caption{Implementation of linear blending.}
    \label{fig:linear_blending}
\end{figure}

Afterward, linear blending was achieved through OpenCV’s \texttt{addWeighted()} function as shown in Fig. \ref{fig:linear_blending}.

\subsection{Version 3 (FCE + CLAHE with Hue Post-processing)}
\begin{figure}[h]
    \centering
    \includegraphics[width=0.9\linewidth]{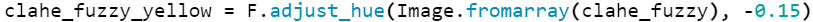}
    \caption{Implementation of hue adjusting.}
    \label{fig:hue_adjusting}
\end{figure}

Post-processing of hue from green to yellow was achieved by the \texttt{adjust\_hue()} function of the Image Python Library as shown in Fig. \ref{fig:hue_adjusting}. As previously detailed, the incorporation of a yellowish hue has proven to be beneficial in aiding doctors in recognizing blood vessels and abnormalities in medical images.

\section{Results and Conclusion}
\label{sec:results}

\begin{figure}[h]
    \centering
    \begin{subfigure}{0.24\linewidth}
        \includegraphics[width=\linewidth]{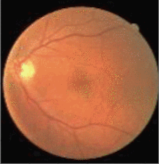}
        \caption{Original}
        \label{fig:res_original}
    \end{subfigure}
    \begin{subfigure}{0.24\linewidth}
        \includegraphics[width=\linewidth]{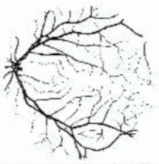}
        \caption{Ground Truth}
        \label{fig:res_segmentation}
    \end{subfigure}
    \begin{subfigure}{0.24\linewidth}
        \includegraphics[width=\linewidth]{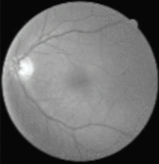}
        \caption{Grayscale}
        \label{fig:res_grayscale}
    \end{subfigure}
    \begin{subfigure}{0.24\linewidth}
        \includegraphics[width=\linewidth]{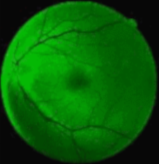}
        \caption{FCE}
        \label{fig:res_fuzzy}
    \end{subfigure}
    
    \begin{subfigure}{0.24\linewidth}
        \includegraphics[width=\linewidth]{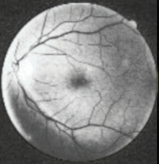}
        \caption{HE}
        \label{fig:res_HE}
    \end{subfigure}
    \begin{subfigure}{0.24\linewidth}
        \includegraphics[width=\linewidth]{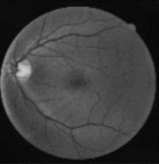}
        \caption{CLAHE}
        \label{fig:res_CLAHE}
    \end{subfigure}
    \begin{subfigure}{0.24\linewidth}
        \includegraphics[width=\linewidth]{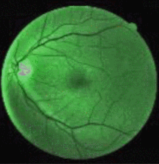}
        \caption{FCE + CLAHE}
        \label{fig:res_fuzzy_clahe}
    \end{subfigure}
    \begin{subfigure}{0.24\linewidth}
        \includegraphics[width=\linewidth]{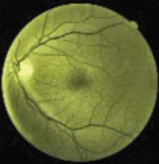}
        \caption{Our Method}
        \label{fig:res_proposed}
    \end{subfigure}

    \caption{Images of the original and contrast-enhanced versions using various methodologies.}
    \label{fig:res_comparison}
\end{figure}

Figure \ref{fig:res_original} presents the original input image, while Figure \ref{fig:res_segmentation} shows the manual segmentation by the experts. The grayscale transformation of the original image is illustrated in Figure \ref{fig:res_grayscale}. The output of Version 1 (FCE, mentioned in Sec. \ref{subsec:version_1}) is displayed in Figure \ref{fig:res_fuzzy}, whereas Figures \ref{fig:res_HE} and \ref{fig:res_CLAHE} depict the results of Histogram Equalization (HE) and CLAHE, two widely used contrast enhancement techniques. Version 2 (FCE + CLAHE, demonstrated in Sec. \ref{subsec:version_2}), shown in Figure \ref{fig:res_fuzzy_clahe}, achieved improved accuracy in enhancing vessels around the macular area. Finally, Version 3 (FCE + CLAHE with Hue Post-processing, explained in Sec. \ref{subsec:version_3}), illustrated in Figure \ref{fig:res_proposed}, further improved contrast visibility by incorporating a yellow image filter.

The output shows that most of the contrast enhancement methods showed visual enhancements compared to the original image. 

\begin{figure}[h]
    \centering
    \includegraphics[width=0.8\linewidth]{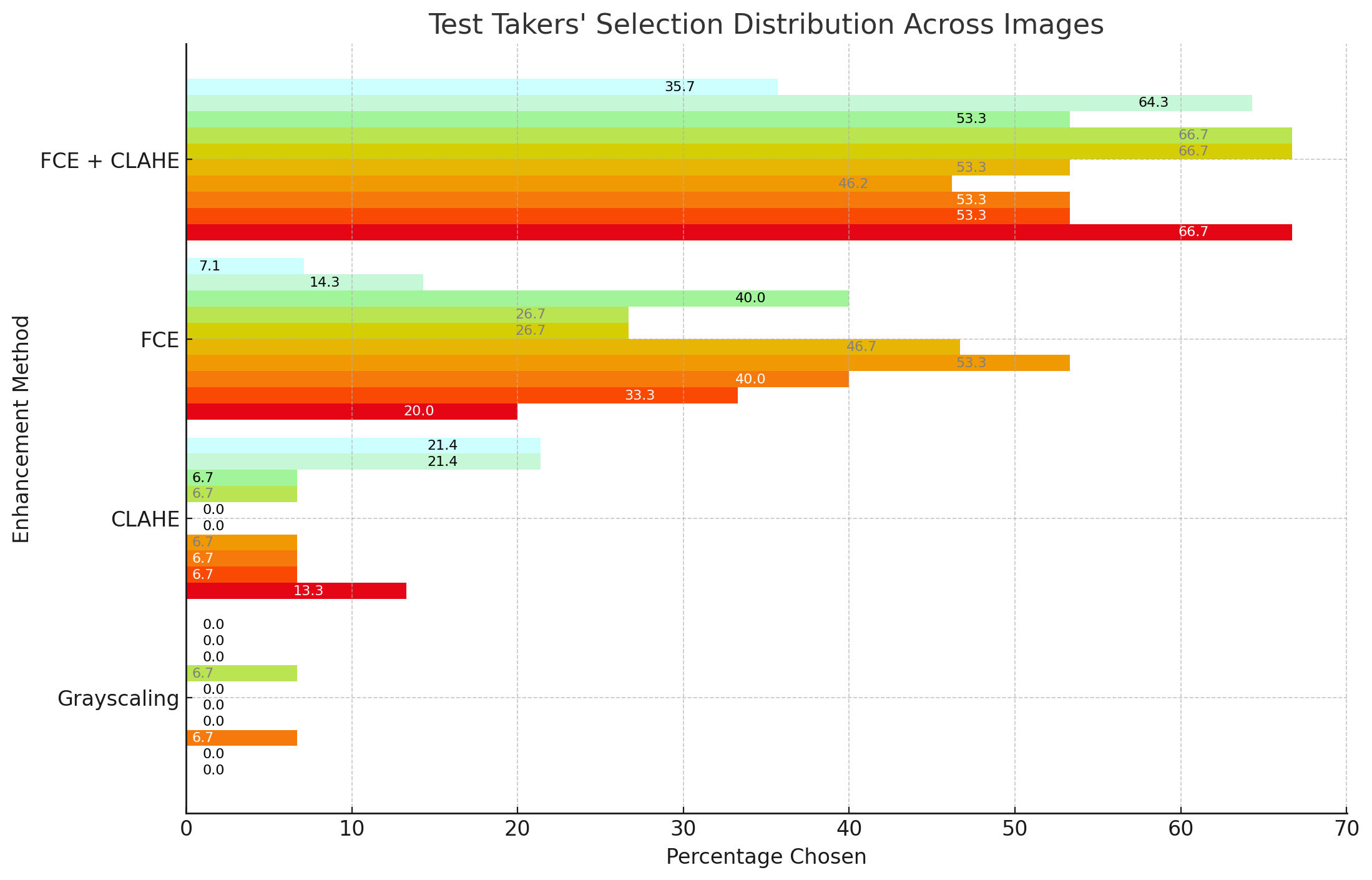}
    \caption{A bar chart displaying how each test taker selected different enhancement methods across multiple images.}
    \label{fig:graph}
\end{figure}

\begin{figure}[h]
    \centering
    \includegraphics[width=0.8\linewidth]{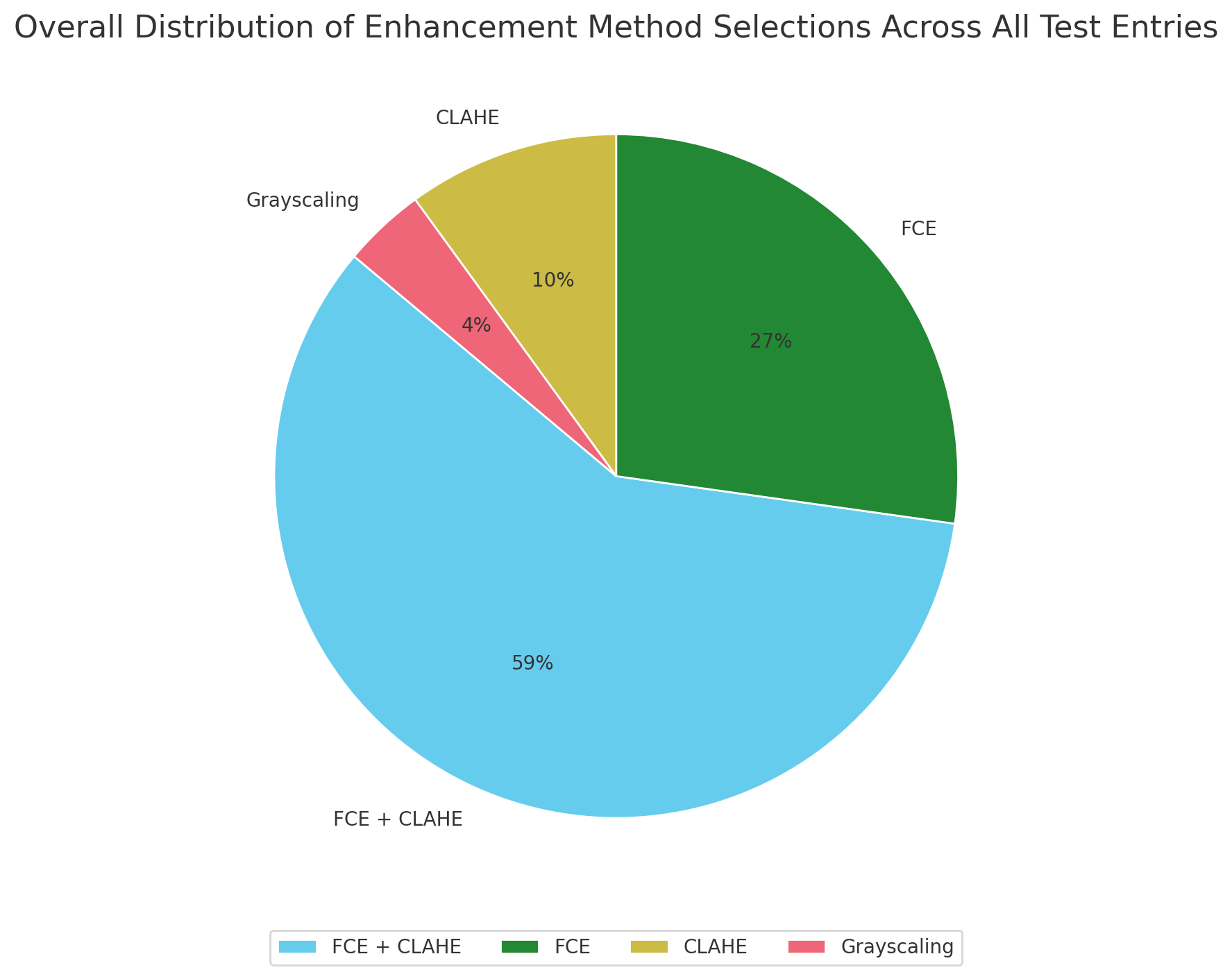}
    \caption{A pie chart representing the total number of selections made for each enhancement method across all test takers and images.}
    \label{fig:pie}
\end{figure}

A survey was conducted using ten individuals to determine which image has the most accurate vessel segmentation. The human survey was done since both the contrast enhancement and the hue modification heavily benefit from human cognition more than that of segmentation algorithms – which utilized their own methods for preprocessing and noisy filtering, tuned for their own needs. Each respondent was provided with manual segmentation of the data by an ophthalmologist and told to choose the best output by comparing each output with its manual segmentation data.  Fig. \ref{fig:graph} and Fig. \ref{fig:pie} show the survey results of different contrast enhancement methods.

Based on the survey results, it has been shown that gray scaling alone is not sufficient for enhancing the visual contrast of retinal images. People tended to prefer CLAHE or gray scaling over Fuzzy Methods for images that had an outlier within the image (bright spots, noises, unhealthy morphology, etc.), whereas pure FCE was preferred over the combination of FCE and CLAHE when the luminosity of the image was evenly distributed.

Nonetheless, it was evident that the combination of the FCE and CLAHE method showed major improvement in general, including improvement over the FCE method. Furthermore, the performance review survey indicates that both the FCE and FCE + CLAHE methods outperformed others, achieving a remarkable 88\% as the preferred enhancement methods. This underscores the effectiveness of preprocessing through fuzzy logic.
\section*{CRediT Author Statement}
\textbf{Adnan Shaout:} Methodology,  Validation, Formal analysis, Investigation, Writing - Review \& Editing, Visualization, Supervision, Project administration, Funding acquisition. \textbf{Jiho Han:} Conceptualization, Methodology, Software, Validation, Formal Analysis, Investigation, Resources, Data curation, Writing - Original Draft, Writing - Review \& Editing, Visualization
\vfill

\bibliographystyle{IEEEtran}
\bibliography{bibliography}

\end{document}